\documentclass{article}

\usepackage[nonatbib,main,preprint]{neurips_2026}

\usepackage[utf8]{inputenc} 
\usepackage[T1]{fontenc}    
\usepackage{hyperref}       
\usepackage{url}            
\usepackage{booktabs}       
\usepackage{amsfonts}       
\usepackage{nicefrac}       
\usepackage{microtype}      
\usepackage[table]{xcolor}         

\usepackage{amsmath}
\usepackage{amssymb}
\usepackage{mathtools}
\usepackage{amsthm}

\usepackage{graphicx}
\usepackage{wrapfig}
\usepackage{subcaption}
\usepackage{multirow}
\usepackage{algorithm}
\usepackage{algpseudocode}

\usepackage{diagbox}
\usepackage{hhline}

\usepackage{siunitx}

\title{AnyDepth-DETR/-YOLO: \\Any-depth object detection with a single network}

\author{%
  Woochul Kang\thanks{Corresponding author} \\
  Incheon Nat'l Univ. \\
  \texttt{wchkang@inu.ac.kr} \\
  \And
  Hyungseop Lee \\
  Incheon Nat'l Univ. \\
  \texttt{hhss0927@inu.ac.kr} \\
    \And
  Jiho Lee \\
  Incheon Nat'l Univ. \\
  \texttt{jiho264@inu.ac.kr} \\
}
\begin{document}

\maketitle

\begin{abstract}
Modern object detectors are static, fixed-depth networks optimized for a
single operating point, requiring separate models for different deployment
scenarios. We present an any-depth detection framework that enables a
single network to span a continuous range of accuracy--efficiency
trade-offs by controlling depth at inference time without retraining.
Each backbone and neck stage is divided into an \emph{essential path},
which always executes, and a skippable \emph{refinement path}; this
decomposition preserves the full multi-scale feature hierarchy at every
depth configuration, unlike conventional early exiting that discards
entire stages. To train such a network, jointly optimizing many
sub-networks of varying depth introduces conflicting gradient signals.
We address this via self-distillation between only the two extremes,
with prediction-level and feature-level alignment losses that enforce
stage-wise modularity, ensuring the outputs of each stage remain
compatible regardless of the paths taken. Instantiated on RT-DETR and
YOLOv12, our full-depth configurations match or surpass their respective
SOTA baselines with negligible parameter overhead, while the most
efficient configurations achieve up to $1.82\times$ speedup 
at a cost of only 2.0 AP, all from a single set of weights.
\end{abstract}    
\section{Introduction}
Object detection underpins applications such as autonomous driving,
video surveillance, and robotics, yet modern detectors achieve high
accuracy at the cost of deep hierarchical architectures that incur
substantial computation and latency. To navigate the accuracy--efficiency
trade-off, practitioners train separate models at different operating
points, each static and optimized for a fixed resource budget, making it
difficult for one network to satisfy heterogeneous deployment requirements.

This rigidity is particularly problematic under varying resource budgets:
a surveillance system may need to process many streams in parallel yet
switch to high-accuracy mode when a scene of interest is detected,
and a cloud service must serve clients with different latency tiers
without maintaining a separate model for each. A single static model
cannot adapt to such fluctuating demands, motivating detectors that
dynamically balance accuracy and efficiency within one network.

Prior adaptive computation methods, including early exiting and width
adaptation, reduce overhead by selectively bypassing layers or channels
at inference, yielding proven efficiency gains in image
classification~\cite{kang2024adaptive, yu2022widthdepth, yu2018slimmable}.
These designs, however, transfer poorly to object detection.
Unlike classifiers that produce a single global output, detectors rely
on a tightly coupled backbone--neck--head pipeline that depends on
hierarchical multi-scale features to localize and classify objects at
varying scales. Naively skipping stages or pruning channels breaks these
dependencies and degrades multi-scale feature quality.
To our knowledge, no prior work has achieved depth-controlled computation
in object detection while preserving the full multi-scale feature
hierarchy that detection requires.

In this work, we present an any-depth architecture and training framework
for multi-scale object detection that enables a single model to span a
continuous range of accuracy--efficiency trade-offs by controlling network
depth. The central design principle is \textit{stage-wise modularity}:
at every stage boundary in the backbone and neck, the output of a lighter
sub-network and that of a fuller sub-network must be interchangeable,
so that downstream stages process either reliably regardless of which
depth configuration is active. Realizing this property requires both
architectural and training solutions that we develop jointly.

\begin{figure}
    \centering
    \includegraphics[width=0.92\linewidth]{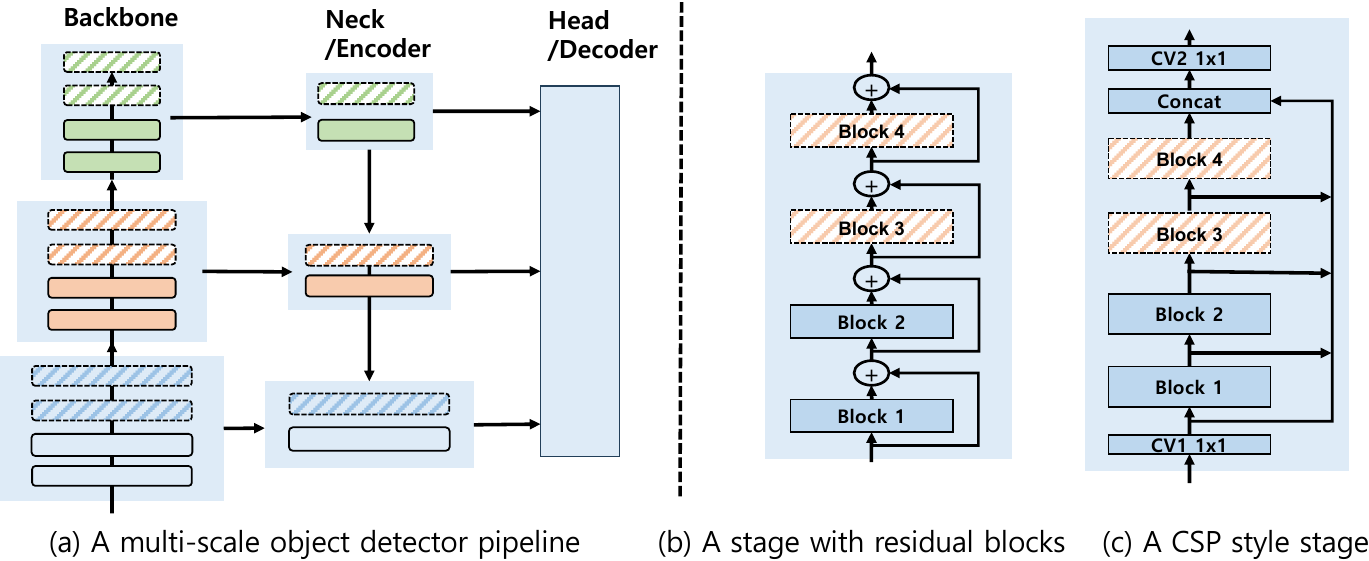}
\caption{(a) Overall pipeline of our any-depth object detection network, following the standard backbone--neck--head design common to modern detectors such as RT-DETR and YOLO, operating on multi-scale features. Each stage in both the backbone and neck is divided into an \emph{essential path} (solid blocks) and a skippable \emph{refinement path} (dashed blocks). At inference time, any combination of essential and full paths across stages yields a valid sub-network, enabling a continuous range of accuracy--efficiency trade-offs from a single set of weights. (b) A residual-block stage, as used in RT-DETR. (c) A CSP-style stage, as used in the YOLO family. In both stage types, the essential and full paths are jointly trained via self-distillation to produce aligned feature representations at each stage boundary, so that downstream stages can process the output of either path reliably.}
    \label{fig:archblocks}
\end{figure}

On the architecture side, as shown in Figure~\ref{fig:archblocks}-(a),
we propose dividing each stage in both the backbone and neck into an
\emph{essential path}, which always executes and learns the core feature
representation at the given scale, and a \emph{refinement path}, which is
skippable and further refines those features when additional computation
is available. Unlike conventional early exiting, which discards entire
stages and breaks the backbone--neck--head coupling, this decomposition
preserves all hierarchical stages at every depth configuration,
structurally enabling stage-wise modularity. Any combination of essential
and full paths across stages yields a valid sub-network, producing a rich
set of depth configurations from a single architecture.

On the training side, naively training all such sub-networks jointly is
computationally prohibitive, and shared weights across paths can introduce
conflicting gradients. We address both by training only the two extremes
(the \emph{super-net} with all full paths and the \emph{base-net} with
all essential paths) jointly via self-distillation. 
The super-net acts as
an online teacher, and prediction-level and feature-level alignment losses
enforce stage-wise modularity through training between the two paths at
every stage boundary. 
Intermediate sub-networks then generalize from the trained
endpoints without requiring direct supervision.

We instantiate our approach on RT-DETR~\cite{zhao2024rtdetr} and
YOLOv12~\cite{tian2025yolov12}, two complementary state-of-the-art
real-time detection paradigms: RT-DETR uses a transformer encoder--decoder
for end-to-end detection, while YOLOv12 follows a CNN backbone with dense
prediction heads and NMS. Although the two architectures differ
substantially, our any-depth framework applies to both;
architecture-specific design choices are described where they arise.
The resulting models, AnyDepth-DETR and AnyDepth-YOLO, are evaluated on
the COCO benchmark. In the full-depth (super-net) configuration, both
models match or surpass their respective state-of-the-art baselines.
At the most efficient depth configuration, AnyDepth-DETR (R-101) achieves
a $1.82\times$ speedup at a cost of only 2.0 AP, and AnyDepth-YOLO (L)
achieves a $1.57\times$ speedup with only 1.8 AP drop. 
Analysis (in Appendix \ref{sec:appendix-analysis}) further shows that the refinement path primarily improves
localization precision rather than detection coverage.
These results
demonstrate that a single modular network can replace multiple fixed-size
detectors, paving the way toward truly adaptive real-time object detection.

\section{Related Work}
\label{sec:related}

\textbf{Real-Time Object Detectors.}
Modern real-time detectors rely on hierarchical multi-scale features aggregated
by a dedicated neck. FPN~\cite{lin2017feature} fuses semantics with spatial
signals via a top-down pathway, and PANet~\cite{liu2018path} adds a bottom-up
shortcut to improve localization. The YOLO family builds on these necks with
NMS-based dense heads; YOLOv12~\cite{tian2025yolov12} is the latest, introducing
area attention. The DETR family~\cite{carion2020detr} replaces the neck with a
transformer encoder and eliminates NMS; DAB-DETR~\cite{liu2022dabdetr},
DN-DETR~\cite{li2022dndetr}, DINO~\cite{zhang2023dino}, and
RT-DETR~\cite{zhao2024rtdetr} progressively refine it with dynamic anchor
queries, denoising, and a redesigned multi-scale encoder. To cover diverse
resource budgets, EfficientDet~\cite{tan2020efficientdet}, Scaled-YOLOv4~\cite{wang2021scaledyolov4},
and EAutoDet~\cite{wang2022eautodet} each train a family of models via compound
scaling, cross-stage partial scaling, and architecture search, respectively.
However, every individual model remains static, 
requiring practitioners to maintain multiple networks. 

\textbf{Adaptive Networks for Object Detection.}
Adaptive inference is well-studied in classification: MSDNet~\cite{huang2018multiscale}
enables anytime prediction via intermediate classifiers, Slimmable
Networks~\cite{yu2018slimmable} support multiple channel widths via switchable
BN, and AdaptiveDepth~\cite{kang2024adaptive} introduces skippable sub-paths
for depth modulation at inference. 
These methods were designed for classification, where a single global representation suffices, making them ill-suited for detection, where backbone, neck, and head are tightly coupled through a multi-scale feature pyramid that must remain intact across all depth configurations.
For detection, DynamicDet~\cite{lin2023dynamicdet} uses a learned routing
module to decide the inference route per image, and
AdaDet~\cite{yang2024adadet} applies bounding-box uncertainty as a per-input
exit criterion. Unlike these input-dependent approaches, our method applies
stage-wise conditional computation across both backbone and neck under a
resource-driven strategy with no routing overhead at inference time.
Work from the real-time computing community has explored scheduling multi-path neural networks under latency constraints across embedded, robotics, and cloud systems~\cite{heo2020realtime,
liu2020onremoving, liu2022self, kang2024qos}, while
AnytimeYOLO~\cite{kuhse2025anytimeyolo} adds early-exits to YOLO for
fixed-granularity interruptible inference. Our models are directly compatible
with such schedulers: depth configurations are selectable at runtime without
retraining, enabling a richer accuracy--efficiency trade-off space than
fixed-exit networks.

\textbf{Knowledge Distillation for Object Detection.}
KD in detection is harder than in classification due to foreground--background
imbalance~\cite{chen2017efficient}; early methods therefore use feature
imitation over foreground regions~\cite{dai2021general, cao2022pkd,
yang2022focal, jia2024mssd}. More recent prediction-level approaches,
Localization Distillation~\cite{zheng2022local} and
CrossKD~\cite{wang2024crosskd}, show that distilling output distributions
can match or exceed feature imitation.
KD-DETR~\cite{wang2024kddetr} further identifies a DETR-specific challenge.
Although several works have explored knowledge distillation for
object detection, KD for state-of-the-art
real-time detectors such as RT-DETR~\cite{zhao2024rtdetr} and
YOLOv12~\cite{tian2025yolov12} remains largely unexplored;
to the best of our knowledge, our work provides the first
systematic evidence that effective self-distillation training
is achievable for these architectures.

\section{Approach}
\subsection{Overview}

\paragraph{Multi-Scale Architecture of Object Detectors.}
As illustrated in Figure~\ref{fig:archblocks}-(a), modern object detectors
consist of a backbone, a neck, and a detection head.
The backbone extracts multi-scale features at progressively coarser
resolutions, the neck fuses them for objects of varying
sizes~\cite{lin2017feature, liu2018path}, and the detection head performs
localization and classification.
Both backbone and neck are organized as a series of stages, each operating at a
specific scale. This multi-scale structure is fundamental to detection:
skipping entire stages removes scale coverage and causes significant
accuracy degradation.

\paragraph{Any-Depth Object Detection.}
To support depth adaptation without sacrificing multi-scale coverage, we
divide each backbone and neck stage into an \emph{essential path}, which
always executes and learns the core representation at the given scale, and
a \emph{refinement path}, which is skippable and further refines those
features when additional computation is available.
Executing both is the \emph{full path}.
The network taking all full paths across stages is the \emph{super-net},
the network taking only essential paths is the \emph{base-net}, and any
combination in between yields a valid intermediate sub-network, all
sharing a single set of weights.
For any such combination to operate correctly, the essential path output
$x_\text{ess}$ and the full path output $x_\text{full}$ at each stage
boundary must be interchangeable, a property we call \emph{stage-wise
modularity}.
This must be both structurally enabled by the architecture
(Section~\ref{sec:arch}) and enforced through training
(Section~\ref{sec:training}), after which any sub-network can be freely
selected at inference from a single set of weights.
Stage-wise modularity is verified quantitatively in
Appendix~\ref{sec:appendix-analysis}: our any-depth detectors maintain
CKA~\cite{CKA} of $0.92$ or above across all stage boundaries, while
non-adaptive baselines degrade to $0.09$--$0.66$.

We instantiate our framework on two complementary real-time detectors,
RT-DETR~\cite{zhao2024rtdetr} and YOLOv12~\cite{tian2025yolov12},
yielding AnyDepth-DETR and AnyDepth-YOLO respectively.

\subsection{Architecture}
\label{sec:arch}

We design the essential and refinement paths for each stage type to
structurally enable stage-wise modularity. The two dominant stage
designs, or residual blocks in RT-DETR and CSP-style aggregation in
YOLO, require different treatments, as described below.

\subsubsection{Stages with Residual Blocks}

ResNet-style stages (Figure~\ref{fig:archblocks}-(b)) are used in the
backbone and neck of RT-DETR. Each block learns a residual function on
top of its input, with the $i$-th block output defined as
\begin{equation}
x^{(i)} = x^{(i-1)} + f^{(i)}(x^{(i-1)}), \quad i = 1, \ldots, S,
\end{equation}
where $x^{(0)}$ is the stage input and $S$ the total number of blocks.
Setting the split point at $m = \lceil S/2 \rceil$, the essential and
full path outputs are
\begin{equation}
\label{eq:residual_stage}
x_{\text{ess}} = x^{(m)}, \qquad
x_{\text{full}} = x^{(S)} = x_{\text{ess}} + \Delta(x_{\text{ess}}),
\end{equation}
where $\Delta(\cdot)$ is the cumulative residual contribution of the
refinement blocks $\{f^{(m+1)}, \ldots, f^{(S)}\}$.
Because both paths share the same tensor shape and the refinement is
purely additive, no architectural conflict arises.
The training objective is to encourage $\Delta(\cdot)$ to learn small
refinements that do not shift the feature distribution seen by downstream
stages, which is enforced by self-distillation in
Section~\ref{sec:training}.
\subsubsection{Stages with CSP-Style Aggregation}
In the YOLO family~\cite{bochkovskiy2020yolov4, wang2023yolov7,
wang2024yolov9, tian2025yolov12}, most stages follow a CSP (Cross Stage
Partial)~\cite{wang2020cspnet} style (Figure~\ref{fig:archblocks}-(c)),
where block outputs are concatenated channel-wise and reduced by an
aggregation layer rather than summed element-wise. This introduces an
architectural conflict absent in residual stages: the concatenated tensor
width depends on how many blocks were executed, so the aggregation weight
cannot be shared between the two paths. A natural remedy is to reuse a
column-sliced submatrix of the full aggregator as the essential path
weight, but this causes conflicting gradients; the full path trains the
shared slice to aggregate $S$ blocks while the essential path trains it
to aggregate only $m$. This conflict is more pronounced when $S$ is large,
as more parameters are subject to conflicting gradients.
We therefore introduce a dedicated \emph{switchable aggregation layer} used only when the refinement path is skipped, fully decoupling its gradient stream from the
full-path aggregator $\mathbf{W}_\text{cv2}$. To limit parameter overhead,
this dedicated weight is applied only to stages with $S \ge 4$ blocks;
for smaller stages ($S < 4$), the gradient conflict is mild and
self-distillation suffices to enforce stage-wise modularity.
The full and essential path outputs are:
\begin{equation}
x_{\text{full}}  = \mathbf{W}_{\text{cv2}}
  \cdot [x^{(0)}, \ldots, x^{(S)}], \qquad
x_{\text{ess}}   = \mathbf{W}_{\text{cv2}}^{\text{ess}}
  \cdot [x^{(0)}, \ldots, x^{(m)}],
\end{equation}
where $\mathbf{W}_{\text{cv2}} \in \mathbb{R}^{c_{\text{out}} \times (S+1)c'}$
and $\mathbf{W}_{\text{cv2}}^{\text{ess}} \in \mathbb{R}^{c_{\text{out}}
\times (m+1)c'}$ are $1{\times}1$ convolutions and $c'$ is the hidden
dimension per block. Stage-wise modularity between $x_{\text{ess}}$ and
$x_{\text{full}}$ is then enforced by self-distillation in
Section~\ref{sec:training}.
\subsubsection{Switchable Batch Normalization}
Blocks in the refinement path observe a single input distribution and
are normalized by a standard BN layer. Blocks in the essential path,
however, serve a dual role: when the essential path is taken alone, its
output must be compatible with the full path output at the stage boundary
to satisfy stage-wise modularity. A single shared BN would normalize the
essential path identically regardless of whether the refinement path is
active, failing to account for the distinct roles the essential path plays
in the two execution modes. Inspired by switchable batch
normalization~\cite{yu2018slimmable, kang2024adaptive}, we equip each
block in the essential path with two independent BN layers: one activated
when the current stage takes the full path, and the other when only the
essential path is taken. 
This allows the essential path to adjust its output distribution
differently in the two modes, promoting compatibility between
$x_{\text{ess}}$ and $x_{\text{full}}$ at every stage boundary while
maintaining normalization accuracy across all depth configurations.


\subsection{Training}
\label{sec:training}

The goal of training is to optimize a single shared-weight network such
that every sub-network, defined by any combination of essential and full
paths across stages, achieves detection performance comparable to a
dedicated model trained solely at that depth configuration.
The key difficulty is that parameter sharing across sub-networks of
different depths introduces conflicting gradient signals: optimizing for
one depth configuration can degrade others.
Our training strategy addresses this through joint supervision of the
super-net and base-net with a \emph{self-distillation} objective that
reduces potential gradient conflicts while promoting stage-wise modularity.

\subsubsection{Self-Distillation Training Objective}

In our self-distillation strategy, the super-net acts as an online teacher
for the base-net within the same shared-parameter network. For each
training iteration, two sequential forward--backward passes are performed
on the same mini-batch.

In the first pass, the super-net executes all full paths and produces
predictions $P^{\text{super}} = (P^{\text{super}}_{\text{cls}},
P^{\text{super}}_{\text{reg}})$ and intermediate feature maps
$\mathcal{F}^{\text{super}}$. The super-net is supervised with the
standard ground-truth detection loss:
\begin{equation}
\mathcal{L}^{\text{super}} =
  \mathcal{L}_{\text{cls}}(P^{\text{super}}_{\text{cls}}, y^{\text{cls}})
+ \mathcal{L}_{\text{reg}}(P^{\text{super}}_{\text{reg}}, y^{\text{reg}}),
\end{equation}
and its gradients are immediately back-propagated. Since the super-net is
identical to the base detector, all training hyperparameters follow the
respective base model without modification.

In the second pass, the base-net executes all essential paths on the same
batch, producing predictions $P^{\text{base}}$ and features
$\mathcal{F}^{\text{base}}$. The base-net loss combines ground-truth
supervision with self-distillation from the super-net:
\begin{equation}
\label{eq:loss_base}
\mathcal{L}^{\text{base}}
    = \alpha\,\mathcal{L}^{\text{gt}}(P^{\text{base}}, y)
    + (1-\alpha)\mathcal{L}^{\text{kd}},
\end{equation}
where $\alpha \in [0, 1]$ balances ground-truth supervision against
self-distillation, and the distillation loss decomposes as:
\begin{equation}
\mathcal{L}^{\text{kd}} = \mathcal{L}^{\text{kd}}_{\text{cls}}
+ \mathcal{L}^{\text{kd}}_{\text{reg}} + \mathcal{L}^{\text{kd}}_{\text{feat}}.
\end{equation}

All distillation terms are computed over the foreground anchors assigned
by the super-net's Task-Aligned Assigner (TAL) for AnyDepth-YOLO, and
the positive queries from the Hungarian matcher for AnyDepth-DETR.
The super-net predictions are treated as a fixed teacher during the
second backward pass.

$\mathcal{L}^{\text{kd}}_{\text{cls}}$ and
$\mathcal{L}^{\text{kd}}_{\text{reg}}$ implicitly promote stage-wise
modularity by driving the base-net's predictions toward those of the
super-net, while $\mathcal{L}^{\text{kd}}_{\text{feat}}$ explicitly
enforces it by aligning essential- and full-path feature representations
at every stage boundary.
Since stage-wise modularity is enforced using only the two extremes,
intermediate sub-networks formed by mixing essential and full paths
generalize without direct supervision, eliminating the need to train all
possible depth configurations explicitly.

\paragraph{Prediction KD.}
For classification, we apply KL divergence over the soft logits:
\begin{equation}
\mathcal{L}^{\text{kd}}_{\text{cls}} =
  \text{KL}(P^{\text{super}}_{\text{cls}} \| P^{\text{base}}_{\text{cls}}).
\end{equation}
We use KL divergence rather than the super-net's classification loss
(BCE for both AnyDepth-YOLO and AnyDepth-DETR) because the super-net's
soft logits encode inter-class relationships --- the ``dark
knowledge''~\cite{hinton2015distilling} --- that BCE would discard, and
it cleanly separates the distillation objective from background
suppression, which is handled solely by the ground-truth loss.

For regression, we use the super-net's predicted boxes as soft targets under the same IoU and edge loss \cite{li2023generalizedfocal} terms as used during training:
\begin{equation}
\mathcal{L}^{\text{kd}}_{\text{reg}} =
  \mathcal{L}_{\text{IoU}}^{\text{kd}}(
    \hat{y}^{\text{super}}_{\text{box}},\,\hat{y}^{\text{base}}_{\text{box}})
+ \mathcal{L}_{\text{edge}}^{\text{kd}}(
    \hat{y}^{\text{super}}_{\text{box}},\,\hat{y}^{\text{base}}_{\text{box}}).
\end{equation}
A key challenge in both terms is the target assignment mismatch between
the super-net and base-net, which we address in
Section~\ref{sec:assignment_conflict}.

\paragraph{Feature Alignment KD.}
\label{sec:feat_kd}
While prediction distillation implicitly encourages stage-wise
modularity, we introduce an explicit feature-level alignment loss
$\mathcal{L}^{\text{kd}}_{\text{feat}}$ that directly enforces
$x_{\text{ess}} \approx x_{\text{full}}$ at each stage boundary,
ensuring any downstream stage receives a compatible input regardless of
which path the preceding stage took.
However, directly matching raw spatial feature maps would
over-constrain the full path, penalizing the spatial refinements that
the refinement path is designed to learn 
(see Appendix~\ref{sec:appendix-analysis},
Table~\ref{tab:feat_alignment_comparison}).
We therefore apply Global
Average Pooling (GAP) channel-wise and align the $\ell_2$-normalized
descriptors across all supervised stages $\mathcal{I}$:
\begin{equation}
\label{eq:feat_kd}
\mathcal{L}^{\text{kd}}_{\text{feat}} =
  \frac{1}{|\mathcal{I}|} \sum_{s \in \mathcal{I}}
  \left\| \hat{\mathbf{f}}^{\text{ess}}_s -
          \hat{\mathbf{f}}^{\text{full}}_s \right\|_2^2,
\qquad
\hat{\mathbf{f}}_s =
  \frac{\text{GAP}(x^{(s)})}{\|\text{GAP}(x^{(s)})\|_2}.
\end{equation}
This is equivalent to $2(1 - \cos(\hat{\mathbf{f}}^{\text{ess}}_s,
\hat{\mathbf{f}}^{\text{full}}_s))$, encouraging directional alignment
rather than value-level imitation. The refinement path thus retains freedom to
develop richer spatial representations, while the aligned channel
descriptors ensure downstream stages can process the output of either
path interchangeably.
As we show in Appendix~\ref{sec:appendix-analysis}, this design
enforces stage-wise modularity: the refinement path improves
localization precision more than detection coverage, and quantitative
analysis confirms high representational similarity across all stage
boundaries.

\begin{figure}
    \centering
    \includegraphics[width=0.9\linewidth]{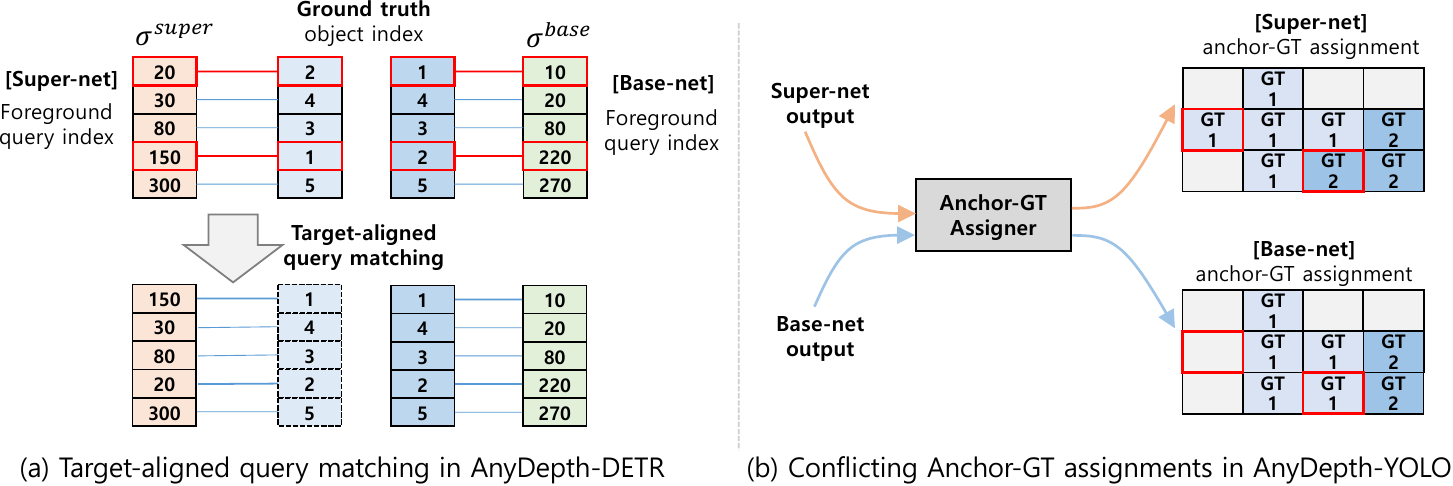}
    \caption{Conflicting target assignment in any-depth detectors.
             (a)~In AnyDepth-DETR, the Hungarian matcher can assign
             different queries to the same ground-truth objects across
             the two execution paths. We reorder the super-net's
             foreground queries to align with the base-net's
             target-matched query ordering before applying KD.
             (b)~In AnyDepth-YOLO, the Task-Aligned Assigner (TAL) can
             assign the super-net and base-net outputs to different
             anchors for the same ground-truth object.
             We exclude conflicting
             anchor assignments (in red boxes) from the foreground distillation set.}
    \label{fig:assignmentconflict}
\end{figure}

\subsubsection{Target Assignment Conflict in Self-Distillation}
\label{sec:assignment_conflict}

A fundamental challenge in self-distillation, where teacher and student
share the same weights and are optimized jointly, is that their
anchor or query assignments to target objects can disagree. Because the
super-net and base-net produce different predictions, the Hungarian
matcher (AnyDepth-DETR) and TAL (AnyDepth-YOLO) may assign the same
anchor or query to different ground-truth objects. Naively applying the
super-net as a teacher on such anchors would inject contradictory
supervision: the base-net would be forced to mimic a prediction that
its own assigner has already allocated to a different target object.

\paragraph{Target-Aligned Query Matching in AnyDepth-DETR.}
RT-DETR assigns predictions to ground-truth objects via bipartite
matching~\cite{kuhn1955hungarian, carion2020detr}, producing a
permutation $\sigma$ that maps each query to at most one ground-truth
object.
As shown in Figure~\ref{fig:assignmentconflict}-(a), the super-net and
base-net solve this matching independently, yielding permutations
$\sigma^{\text{super}}$ and $\sigma^{\text{base}}$ that may assign the
same ground-truth object to different query positions.
At query position $k$, the base-net predicts an output for object
$\sigma^{\text{base}}(k)$, while the super-net predicts an output for
object $\sigma^{\text{super}}(k)$. When
$\sigma^{\text{super}}(k) \neq \sigma^{\text{base}}(k)$, element-wise
distillation forces the base-net to mimic the teacher's response for a
\emph{different} object, introducing a spurious supervisory signal.
To avoid this, we re-order the super-net's matched predictions before
computing the KD loss so that both networks respond to the same
ground-truth object at every position, ensuring semantically consistent
per-object guidance.

\paragraph{Conflicting Assignment Removal in AnyDepth-YOLO.}
Let $\mathcal{A}^{\text{super}}$ and $\mathcal{A}^{\text{base}}$ be
the sets of foreground anchor indices assigned by TAL for the super-net
and base-net respectively, and let $g^{\text{super}}_j$ and
$g^{\text{base}}_j$ denote the ground-truth index each assigns to
anchor $j$.
As shown in Figure~\ref{fig:assignmentconflict}-(b), the two
assignments can conflict. We define the \emph{conflict set}:
\begin{equation}
\mathcal{C} = \bigl\{\, j \in \mathcal{A}^{\text{super}}
  \cap \mathcal{A}^{\text{base}} \;\bigm|\;
  g^{\text{super}}_j \neq g^{\text{base}}_j \,\bigr\},
\end{equation}
i.e., anchors that are foreground in both sub-networks but point to
different objects. KD losses are then restricted to the
\emph{valid distillation set}:
\begin{equation}
\mathcal{V} = \bigl(\mathcal{A}^{\text{super}}
  \cap \mathcal{A}^{\text{base}}\bigr) \setminus \mathcal{C},
\end{equation}
which contains only anchors that are jointly foreground in both
sub-networks and agree on the target object, providing
unambiguous and stable regression targets for distillation.
The remaining anchors, including background anchors, are supervised solely by the ground-truth
loss in Eq.~\ref{eq:loss_base}.


\begin{table}[t]
  \centering
  \caption{Comparison with SOTA real-time detectors on COCO val2017. 
  FPS is measured at batch size 1 on a single RTX 4090 GPU.}
  \label{tab:sota}
  \small
  \setlength{\tabcolsep}{5pt}
  \renewcommand{\arraystretch}{1.1}
  \begin{tabular}{l l l r r r r r}
    \toprule
    \textbf{Method} & \textbf{Backbone} &
    \textbf{Params (M)} & \textbf{GFLOPs} &
    \textbf{FPS} & \textbf{AP} & \textbf{AP$_{50}$} & \textbf{AP$_{75}$}\\
    \midrule
    \multirow{4}{*}{RT-DETR~\cite{zhao2024rtdetr}}
      & R-101 & 76   & 257 & 78.6  & 54.3          & 72.7  & 58.8\\
      & R-50  & 42   & 134 & 113.4 & 53.1          & 71.3 & 57.7\\
      & R-34  & 31   &  92 & 163.9 & 48.9          & 66.8 & 52.6 \\
      & R-18  & 20   &  60 & 217.4 & 46.5          & 63.8 & 50.3\\[4pt]
      \cline{2-8}
    \multirow{2}{*}{AnyDepth-DETR (R-101)}
      & super-net & 76(+0.065) & 257 &  78.6 & \textbf{54.6} & 72.4 & 59.1\\
      & base-net  &  --  & 141 & 125.6 & 52.6          & 70.8 & 57.1\\[4pt]
      \cline{2-8}
    \multirow{2}{*}{AnyDepth-DETR (R-50)}
      & super-net & 42(+0.037) & 134 & 113.4 & \textbf{53.3} & 71.4 & 57.7\\
      & base-net  &  --  &  79 & 165.3 & 50.6          & 68.7 & 54.8\\
    \midrule
    \multirow{3}{*}{YOLOv12~\cite{tian2025yolov12}}
      & X & 59.4 & 185.9 &  85.5 & 55.4          & 72.6 & 60.3\\
      & L & 26.4 &  83.3 &  101.5 & 53.8          & 71.1 & 58.6\\
      & M & 20.2 &  60.4 & 159.2 & 52.5          & 70.0 & 57.1\\[4pt]
      \cline{2-8}
    \multirow{2}{*}{AnyDepth-YOLO (X)}
      & super-net & 61.3(+1.9) & 185.9 &  85.5 &  \textbf{55.5}  &  71.9 & 60.2 \\
      & base-net  &  --  & 132.9 & 131.8 &   54.0   &  70.4 & 58.6 \\[4pt]
      \cline{2-8}
    \multirow{2}{*}{AnyDepth-YOLO (L)}
      & super-net & 27.3(+0.9) &  83.3 &  101.5 & \textbf{53.9} & 70.4 & 58.6\\
      & base-net  &  --  &  59.6 & 155.3 & 52.1          & 68.7 & 56.6\\
    \bottomrule
  \end{tabular}
\end{table}

\section{Experiments}
\label{sec:exp}

\subsection{Experimental Setup}
\label{sec:setup}
\begin{wrapfigure}{r}{0.41\linewidth}
  \centering
  \vspace{-\baselineskip}  
  \includegraphics[width=\linewidth]{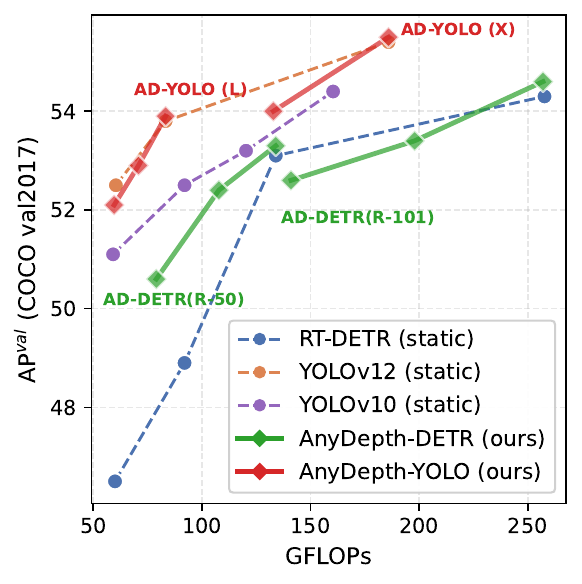}
  \caption{Pareto frontier on COCO.
           }
  \label{fig:pareto}
\vspace{-12mm}  
\end{wrapfigure}
All models are trained and evaluated on the COCO~\cite{lin2015coco} benchmark
(\texttt{train2017} / \texttt{val2017}). 
We adopt the original training recipe of each base detector without modification:
AnyDepth-YOLO follows YOLOv12~\cite{tian2025yolov12} and AnyDepth-DETR follows
RT-DETR~\cite{zhao2024rtdetr} in all settings, including learning rate schedule,
data augmentation, batch size, and training epochs.
The additional hyperparameters for the self-distillation 
are detailed in Appendix~\ref{sec:appendix-training}.

\subsection{Comparison with State-of-the-Art}
\label{sec:sota}
Table~\ref{tab:sota} reports AP, GFLOPs, and FPS for each model,
and Figure~\ref{fig:pareto} visualizes the accuracy--efficiency Pareto frontier.
Our super-net configurations match or surpass their respective baselines with
negligible parameter overhead: switchable batch normalization adds less than
0.09\% extra parameters, while the switchable aggregator in AnyDepth-YOLO
incurs approximately 3\% additional parameters.
Despite this minimal overhead, the base-net configurations achieve
substantially higher throughput at a modest AP cost, and the continuous
range of intermediate configurations spans the accuracy--efficiency trade-off
covered by multiple individually trained static models.

This flexibility stands in sharp contrast to the static baselines.
For RT-DETR, four separately trained models are required to cover a range
from 60 GFLOPs / 20M parameters (R-18) to 259 GFLOPs / 76M parameters (R-101).
AnyDepth-DETRs (R-50 and R-101), by contrast, support up to 48\% reduction in GFLOPs instantly at inference time without any retraining.
Together, they cover the full accuracy--efficiency range of the RT-DETR family.

\subsection{Performance Across Depth Configurations}
\label{sec:ablation_skippable}

A key advantage of the proposed any-depth detectors is that, once trained,
the depth of each component can be adjusted on the fly at inference time
without any retraining. To characterize the resulting accuracy--efficiency
trade-offs, we sweep each depth dimension of AnyDepth-DETR (R-50) in
Table~\ref{tab:adn_ablation}. Note that decoder early-exiting is an
inherent property of DETR-family models~\cite{carion2020detr}, where
queries are progressively refined across decoder layers, making any
intermediate layer a valid exit point.

Among the skippable components, reducing the encoder depth trades 0.9 AP
for a modest 6\% FPS gain, while routing all backbone stages through the
essential path costs a further 2.0 AP with a 15\% FPS improvement.
Decoder depth 4 offers the best efficiency trade-off: it achieves 53.1 AP
while improving throughput by 11\%. Overall, the base-net achieves a
$1.46\times$ speedup over the super-net at the cost of 2.7 AP, all from a single
set of weights. Notably, although only the two extremes (the super-net
and base-net) are trained explicitly, these intermediate sub-networks
generalize well, offering smooth AP--efficiency trade-offs across the full
depth spectrum. 
This generalization is underpinned by stage-wise modularity, which is
verified quantitatively in Appendix~\ref{sec:appendix-analysis}.

\begin{table}[h]
\centering
\caption{Component-wise ablation of AnyDepth-DETR (R-50) on COCO
         \texttt{val2017}. Each group varies one component while keeping
         all others at the super-net setting. FPS measured at batch size 1
         on a single RTX 4090 GPU. See Table~\ref{tab:adn_yolo_ablation}
         for the corresponding AnyDepth-YOLO (L) results.}
\label{tab:adn_ablation}
\small
\setlength{\tabcolsep}{6pt}
\renewcommand{\arraystretch}{1.15}
\begin{tabular}{l l r r c}
\toprule
\textbf{Component} & \textbf{Setting} & \textbf{GFLOPs} & \textbf{FPS} & \textbf{AP} \\
\midrule
\rowcolor{gray!12}
\multicolumn{2}{l}{\textbf{Super-net}: all full paths, decoder depth 6}
                                              & 134.2 & 113.2 & \textbf{53.3} \\
\midrule
\multirow{4}{*}{\shortstack[l]{Backbone\\{\small(4 stages)}}}
 & Ess.\ path in P2  & 130.6 & 114.9 & 52.0 \\
 & Ess.\ paths in P2, P3 & 123.4 & 119.6 & 52.0 \\
 & Ess.\ paths in P2, P3, P4 & 112.8 & 127.3 & 51.4 \\
 & Ess.\ paths in P2, P3, P4, P5 & 109.2 & 130.3 & 51.3 \\
\midrule
\multirow{2}{*}{\shortstack[l]{Encoder\\{\small(FPN + PAN)}}}
 & Ess.\ path in PAN         & 128.9 & 116.5 & 53.1 \\
 & Ess.\ paths in FPN + PAN   & 107.8 & 120.2 & 52.4 \\
\midrule
\multirow{5}{*}{\shortstack[l]{Decoder\\{\small(6 layers)}}}
 & 5 layers & 132.2 & 118.9 & 53.2 \\
 & 4 layers & 130.4 & 126.9 & 53.1 \\
 & 3 layers & 128.4 & 136.8 & 52.6 \\
 & 2 layers & 126.4 & 148.0 & 51.3 \\
 & 1 layer  & 124.6 & 160.0 & 48.8 \\
\midrule
\rowcolor{gray!12}
\multicolumn{2}{l}{\textbf{Base-net}: all ess. paths, decoder depth 4}
                                              &  79.0 & 165.3 & \textbf{50.6} \\
\bottomrule
\end{tabular}
\end{table}

\subsection{Ablation}
Table~\ref{tab:ablation_components} ablates each proposed architectural and
training component, starting from a baseline that supervises both
sub-networks with ground-truth only ($\alpha=1$ in Eq.~\ref{eq:loss_base}).
The baseline already achieves competitive performance, demonstrating that
preserving all multi-scale stages at every depth configuration through the
essential and refinement path decomposition is itself a strong inductive
bias for any-depth object detection.

\paragraph{Switchable BN and Aggregator.}
Switchable BN improves both models by decoupling the running statistics of
the two execution paths. The gain is larger for AD-YOLO
(AP$_\text{base}$: 50.8 $\to$ 51.6) than AD-DETR (50.0 $\to$ 50.1), as
CSP-style concat stages produce more statistically distinct distributions
than the additive residual stages of RT-DETR, making BN decoupling
especially important. For AD-YOLO, the dedicated switchable aggregator
$\mathbf{W}^\text{ess}_\text{cv2}$ further improves AP$_\text{super}$
from 53.3 to 53.6 and AP$_\text{base}$ from 51.6 to 52.0, confirming
that fully decoupling the gradient streams of the two aggregation weights
is essential for CSP-style stages.

\paragraph{Self-Distillation and Target Alignment.}
Naive self-distillation alone does not improve performance, indicating
that uncorrected target assignment conflicts inject noisy gradients when
distillation is applied without alignment. Applying target-aligned query
reordering (AD-DETR) and conflicting assignment removal (AD-YOLO) resolves
this: AD-DETR AP$_\text{super}$ recovers to 52.9 and AP$_\text{base}$
improves to 50.3, while AD-YOLO gains only $+$0.1 AP in both super-net and base-net.
The smaller gain in AD-YOLO is expected: anchor-target misassignment in
dense prediction is rare (${<}$1.0\% of anchors), whereas RT-DETR assigns
predictions from only 300 queries, making conflicting matches across
sub-networks proportionally more disruptive.

\paragraph{Feature Alignment.}
Finally, adding $\mathcal{L}^\text{kd}_\text{feat}$ (Eq.~\ref{eq:feat_kd})
to enforce stage-wise modularity at every stage boundary further improves
AD-DETR to AP$_\text{super}$ 53.3 and AP$_\text{base}$ 50.6, while
AD-YOLO remains unchanged. The difference is explained by the stage
architecture. In CSP-style stages, the switchable aggregator
$\mathbf{W}^\text{ess}_\text{cv2}$ already provides structural decoupling
between the two paths, making $\mathcal{L}^\text{kd}_\text{feat}$
redundant: the loss remains negligibly small (${\approx}0.005$) even
without the explicit alignment term, indicating that stage-wise modularity
is already achieved by the architecture alone. In residual stages, by
contrast, the refinement in skippable blocks is purely additive and no
such structural decoupling exists, so the explicit alignment loss is
necessary to enforce stage-wise modularity through training.

\begin{table}[t]
\centering
\caption{Ablation of architecture and training components on COCO
  \texttt{val2017}. AP$_\text{super}$ and AP$_\text{base}$ denote
  AP of the super-net and base-net, respectively.}
\label{tab:ablation_components}
\small
\setlength{\tabcolsep}{4pt}
\renewcommand{\arraystretch}{1.15}
\begin{tabular}{l ccc c cc cc}
\toprule
& \multicolumn{3}{c}{\textbf{Component}} & &
  \multicolumn{2}{c}{\textbf{AD-DETR (R-50)}} &
  \multicolumn{2}{c}{\textbf{AD-YOLO (L)}} \\
\cmidrule(lr){2-4} \cmidrule(lr){6-7} \cmidrule(lr){8-9}
\textbf{Setting} &
  \shortstack{Switch.\\BN} &
  \shortstack{Switch.\\Agg.} &
  \shortstack{Target\\Alig.} &
  \shortstack{Feature\\Alig.} &
  AP$_\text{super}$ & AP$_\text{base}$ &
  AP$_\text{super}$ & AP$_\text{base}$ \\
\midrule
Baseline (naive joint-train) &  &   &  &  & 52.8 & 50.0 & 50.2& 50.8\\
+ Switch.\ BN      & \checkmark &    &   &  & 53.0& 50.1 & 53.3 & 51.6 \\
+ Switch.\ Aggr.   & \checkmark & \checkmark &   &    & -- & -- & 53.6& 52.0\\
\midrule
+ Self-distillation (naive)  & \checkmark & \checkmark &   &   & 52.5 & 50.1 & 53.8 & 52.0 \\
+ Target Align.    & \checkmark & \checkmark & \checkmark &            & 52.9 & 50.3 &53.9 & 52.1\\
+ Feature Align. (Ours)   & \checkmark & \checkmark & \checkmark & \checkmark & 53.3& 50.6 & 53.9& 52.1\\
\bottomrule
\end{tabular}
\end{table}

\section{Conclusions}
\label{sec:conc}
We presented an any-depth object detection framework that enables a single
network to span a continuous range of accuracy--efficiency trade-offs by
controlling depth at inference time. By partitioning each backbone and neck
stage into an essential path and a skippable refinement path, our method
preserves the full multi-scale feature hierarchy at every depth
configuration, providing a practical solution for deploying detectors under
dynamic resource constraints. The central design principle is stage-wise modularity,
a property jointly achieved by the architectural decomposition and
self-distillation training, which ensures that the output of any stage
remains compatible regardless of which paths are taken. Experiments on COCO
demonstrate that a single model can replace multiple fixed-size detectors
without retraining, significantly reducing development and deployment costs.

\paragraph{Limitations.}
(1) Self-distillation training procedure requires two sequential forward-backward passes per iteration, which effectively doubles the training time compared to standard models. (2) The current framework selects depth configurations based on a resource
budget rather than adapting to input difficulty. Integrating a lightweight
routing mechanism for input-adaptive depth selection remains an interesting
direction for future work.



\bibliography{main}
\bibliographystyle{nips}

\clearpage


\appendix
\section{Appendix: Detailed Settings, Analysis, and Evaluation Results}

\subsection{Implementation Details}
\label{sec:appendix_impl}

\paragraph{Architecture Configuration.}
Table~\ref{tab:arch_config} summarizes the per-stage architecture of
AnyDepth-DETR (R-50) and AnyDepth-YOLO (L), showing the total number of
blocks per stage, the essential-path split point $m = \lceil S/2 \rceil$,
and the number of skippable refinement blocks. 
For AnyDepth-DETR, the
backbone follows ResNet-50 with Bottleneck blocks; for AnyDepth-YOLO, the
backbone and neck follow YOLOv12-L with C3k2 and A2C2f blocks. In both
models, the essential path always executes and the refinement path is
skippable at inference time. One exception to the split point formula is
encoder stages in AnyDepth-DETR with $S=3$ blocks, where we use $m=1$
instead of $m=2$, as this configuration yields better empirical performance.

\begin{table}[h]
\centering
\caption{Per-stage architecture configuration of AnyDepth-DETR (R-50) and
         AnyDepth-YOLO (L). $S$: total blocks per stage;
         $m$: essential-path blocks; $S{-}m$: skippable blocks.
         $^*$Stages with $S \ge 4$ use a dedicated switchable aggregation
         layer $\mathbf{W}^\text{ess}_\text{cv2}$.}
\label{tab:arch_config}

\begin{subtable}{\linewidth}
\centering
\caption{AnyDepth-DETR (R-50)}
\begin{tabular}{llcccc}
\toprule
\textbf{Component} & \textbf{Stage} & \textbf{Block type} &
$S$ & $m$ & $S{-}m$ \\
\midrule
\multirow{4}{*}{Backbone}
  & P2 & Bottleneck & 3 & 2 & 1 \\
  & P3 & Bottleneck & 4 & 2 & 2 \\
  & P4 & Bottleneck & 6 & 3 & 3 \\
  & P5 & Bottleneck & 3 & 2 & 1 \\
\midrule
\multirow{2}{*}{Encoder}
  & P4 (FPN)         & RepVggBlock & 3 & 1 & 2 \\
  & P3 (FPN)         & RepVggBlock & 3 & 1 & 2 \\
  & P4 (PAN)         & RepVggBlock & 3 & 1 & 2 \\
  & P5 (PAN)         & RepVggBlock & 3 & 1& 2 \\
\midrule
Decoder & 6 layers & Transformer
        & 6 & \multicolumn{2}{c}{Any layer is a valid exit} \\
\bottomrule
\end{tabular}
\end{subtable}

\vspace{1em}

\begin{subtable}{\linewidth}
\centering
\caption{AnyDepth-YOLO (L)}
\begin{tabular}{llcccc}
\toprule
\textbf{Component} & \textbf{Stage} & \textbf{Block type} &
$S$ & $m$ & $S{-}m$ \\
\midrule
\multirow{4}{*}{Backbone}
  & P2 & C3k2     & 2 & 1 & 1 \\
  & P3 & C3k2     & 2 & 1 & 1 \\
  & P4 & A2C2f$^*$ & 4 & 2 & 2 \\
  & P5 & A2C2f$^*$ & 4 & 2 & 2 \\
\midrule
\multirow{2}{*}{Neck}
  & P4 (FPN) & A2C2f & 2 & 1 & 1 \\
  & P3 (FPN) & A2C2f & 2 & 1 & 1 \\
  & P4 (PAN) & A2C2f & 2 & 1 & 1 \\
  & P5 (PAN) & C3k2 & 2 & 1 & 1 \\
\midrule
Detector head & \multicolumn{5}{l}{No depth adaptation} \\
\bottomrule
\end{tabular}
\end{subtable}

\end{table}

\paragraph{Switchable Aggregation in AnyDepth-YOLO.}
Each CSP-style stage in AnyDepth-YOLO contains a dedicated switchable
aggregation layer $\mathbf{W}^\text{ess}_\text{cv2}$ (Eq.~3) that is
used only when the refinement path is skipped,
fully decoupling its gradient stream from the full-path aggregator
$\mathbf{W}_\text{cv2}$.
To limit parameter overhead, this dedicated weight is applied only to
stages with $S \ge 4$ blocks (marked $^*$ in Table~\ref{tab:arch_config}(b)).

\subsection{Training Details}
\label{sec:appendix-training}

\paragraph{Training Environment.}
All models are trained using PyTorch with
Distributed Data Parallel (DDP) across 4 RTX 4090 GPUs or 4 RTX 5090 GPUs.
All experiments use CUDA 12.8 with automatic mixed precision (AMP)
to reduce memory usage and accelerate training.
FPS is measured separately on a single NVIDIA RTX 4090 GPU at batch size~1,
with 200 warm-up iterations prior to timing.

\begin{table}[htp]
\centering
\caption{Self-distillation hyperparameters for AnyDepth-DETR (R-50) and
         AnyDepth-YOLO (L).}
\label{tab:hyper_selfdistill}
\begin{tabular}{lcc}
\toprule
\textbf{Hyperparameter} & \textbf{AnyDepth-DETR} & \textbf{AnyDepth-YOLO} \\
\midrule
$\alpha$ (Eq.~\ref{eq:loss_base})
    & 0.0   & 0.2 \\
\midrule
\multicolumn{3}{l}{\textit{Classification KD ($\mathcal{L}_\text{cls}^\text{kd}$)}} \\
Weight
    & 3.75  & 0.4 \\
KL divergence temperature
    & 1.0   & 2.0 \\
\midrule
\multicolumn{3}{l}{\textit{Regression KD ($\mathcal{L}_\text{reg}^\text{kd}$)}} \\
IoU loss weight ($\mathcal{L}_\text{IoU}^\text{kd}$)
    & 2.0   & 1.2 \\
Edge loss weight ($\mathcal{L}_\text{edge}^\text{kd}$)
    & 5.0   & 0.8 \\
Edge loss type
    & L1    & DFL\\
KD divergence temperature for DFL
    & --    & 1.0 \\
\midrule
\multicolumn{3}{l}{\textit{Feature Alignment KD ($\mathcal{L}_\text{feat}^\text{kd}$)}} \\
Backbone stages weight
    & 0.5   & 0.4 \\
Encoder stages weight
    & 0.2   & 0.4 \\
\bottomrule
\end{tabular}
\end{table}

\paragraph{Hyperparameters.}
Both AnyDepth-YOLO and AnyDepth-DETR adopt the original training recipe of their
respective base detectors without modification, following YOLOv12~\cite{tian2025yolov12}
and RT-DETR~\cite{zhao2024rtdetr} respectively.
For self-distillation, the super-net is supervised by ground-truth labels
following each base detector's training recipe, while the base-net is jointly
supervised by both ground-truth labels and the super-net's predictions
(Eq.~\ref{eq:loss_base}).
Table~\ref{tab:hyper_selfdistill} lists the hyperparameters specific to
self-distillation.

For AnyDepth-DETR, ResNet backbones are pre-trained before fine-tuning on COCO,
following RT-DETR training recipe \cite{paddleclas2024distillation}.
The super-net path of the backbone is supervised by a
ResNeXt-101 teacher~\cite{yalniz2019billion}, and the base-net path
is trained to match its representations via self-distillation.

\subsection{Analysis}
\label{sec:appendix-analysis}

To better understand how our any-depth detectors achieve their accuracy--efficiency
trade-offs, we analyze two complementary aspects of the trained models:
how the refinement path contributes to detection quality, and whether
stage-wise modularity is empirically achieved across all stage boundaries.

\paragraph{Precision and Recall Analysis.}

\begin{table}[t]
\centering
\caption{Detailed COCO \texttt{val2017} evaluation results for AnyDepth-DETR (R-50),
         comparing the base-net and super-net.
         $\Delta$ denotes the gain of the super-net over the base-net.
         The AP gain consistently exceeds the AR gain across all scales and
         IoU thresholds, indicating that the refinement path primarily improves
         localization precision rather than detection coverage.}
\label{tab:detailed_r50}
\small
\setlength{\tabcolsep}{5pt}
\renewcommand{\arraystretch}{1.15}
\begin{minipage}{0.48\linewidth}
\centering
\begin{tabular}{l ccc}
\toprule
\textbf{Avg. precision} & \textbf{Base-net} & \textbf{Super-net} & $\boldsymbol{+\Delta}$ \\
\midrule
All (0.50:0.95) & 50.6 & 53.3 & $+$2.7 \\
All (0.50)      & 68.7 & 71.4 & $+$2.7 \\
All (0.75)      & 54.8 & 57.7 & $+$2.9 \\
AP$_\text{small}$  & 32.6 & 34.9 & $+$2.3 \\
AP$_\text{medium}$ & 55.0 & 57.6 & $+$2.6 \\
AP$_\text{large}$  & 66.8 & 70.4 & $+$3.6 \\
\bottomrule
\end{tabular}
\end{minipage}
\hfill
\begin{minipage}{0.48\linewidth}
\centering
\begin{tabular}{l ccc}
\toprule
\textbf{Avg. recall} & \textbf{Base-net} & \textbf{Super-net} & $\boldsymbol{+\Delta}$ \\
\midrule
AR$_1$             & 38.1 & 39.2 & $+$1.1 \\
AR$_{10}$          & 64.2 & 65.9 & $+$1.7 \\
AR$_{100}$         & 70.8 & 72.7 & $+$1.9 \\
AR$_\text{small}$  & 53.9 & 56.0 & $+$2.1 \\
AR$_\text{medium}$ & 74.9 & 76.9 & $+$2.0 \\
AR$_\text{large}$  & 87.1 & 88.3 & $+$1.2 \\
\bottomrule
\end{tabular}
\end{minipage}
\end{table}

\begin{figure}[t]
    \centering
    \includegraphics[width=1.0\linewidth]{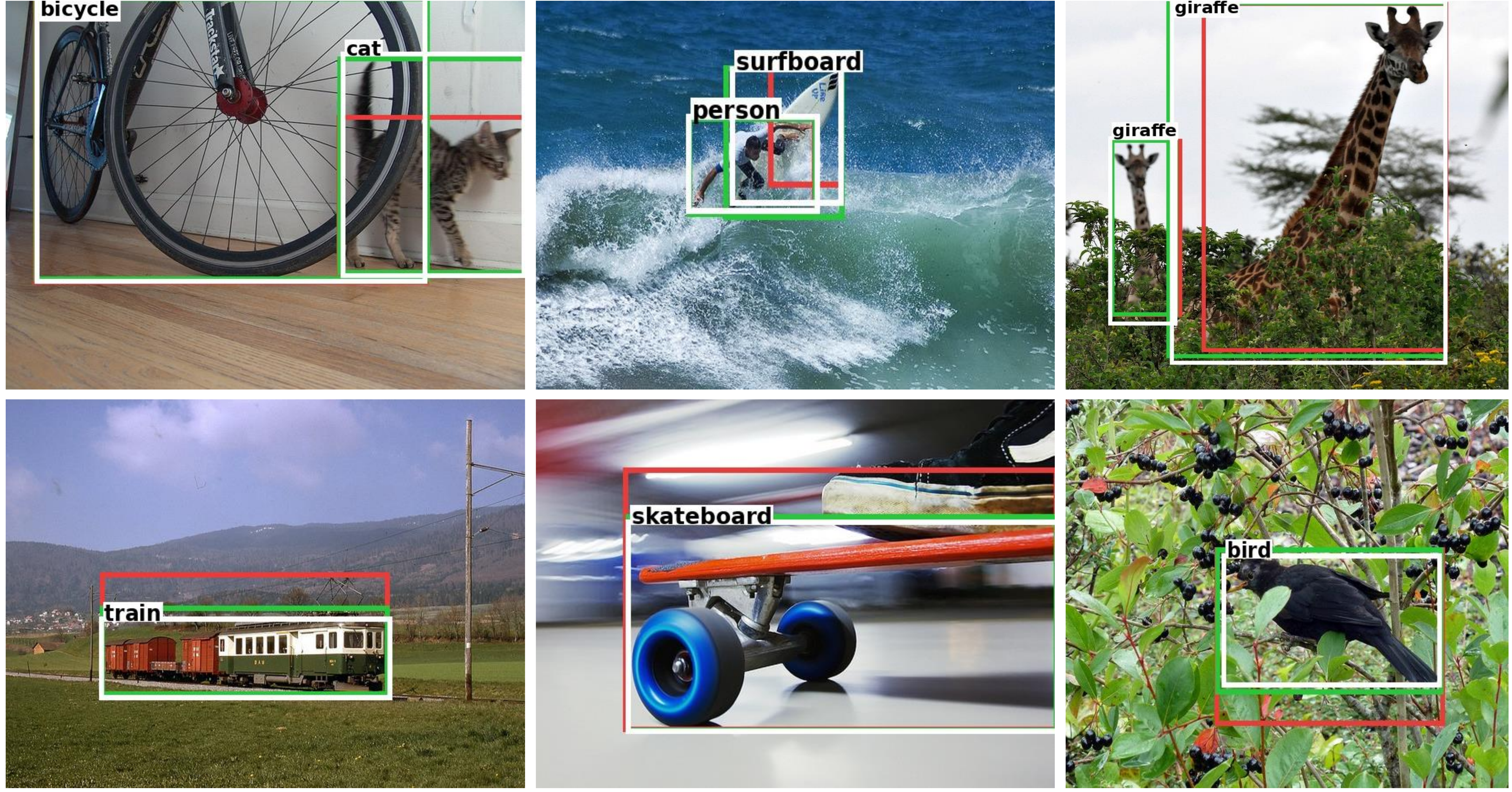}
\caption{AnyDepth-DETR (R-50) localization examples on COCO \texttt{val2017}.
         White: ground truth, Green: super-net, Red: base-net.
         Both sub-networks detect the same objects, but the super-net
         consistently produces tighter bounding boxes on hard images.
         When scenes contain occlusion or clutter (e.g., cat and surfboard),
         the skippable refinement path further refines the basic representation
         from the base-net, recovering finer spatial detail that strict IoU
         thresholds reward.}
    \label{fig:betterlocalization}
\end{figure}
Table~\ref{tab:detailed_r50} breaks down the super-net and base-net results of AnyDepth-DETR(R-50)
on COCO \texttt{val2017}
by IoU threshold and object scale.
The overall AP gain from the refinement path is $+$2.7, and the
strict-IoU gain (AP$_{0.75}$: $+$2.9) slightly exceeds the loose-IoU
gain (AP$_{0.50}$: $+$2.7), indicating that the refinement path sharpens
localization \emph{precision} more than detection coverage.
Consistent with this, the AR gain is smaller than the AP gain across all
scales (AR$_{100}$: $+$1.9 vs.\ AP: $+$2.7), suggesting that the base-net
already recalls most objects reliably, meaning the semantic recognition
signal is largely intact without the refinement path, but localizes them
less precisely.
The refinement path therefore primarily contributes spatial boundary
refinement rather than semantic recognition: it further processes the
base-net's representation to recover finer spatial detail that strict
IoU thresholds reward.

Figure~\ref{fig:betterlocalization} illustrates this qualitatively:
the super-net (green) consistently produces tighter bounding boxes than
the base-net (red), particularly for hard images containing occlusion
and clutter, while both networks detect the same objects.

\paragraph{Stage-Wise Modularity.}
To quantitatively verify stage-wise modularity, we measure Centered
Kernel Alignment (CKA)~\cite{CKA} between the essential and full path
outputs at each stage boundary on COCO \texttt{val2017}. CKA measures
representational similarity independently of rotation or scaling (1.0 =
identical); high CKA at every stage boundary confirms that downstream
stages receive compatible inputs regardless of which path the preceding
stage took. As references, we measure (1) the naive joint-train baseline
from Table~\ref{tab:ablation_components} (our architecture, $\alpha=1$,
no switchable components or distillation), and (2) standard RT-DETR and
YOLOv12, applying the same split point post-hoc in both cases.
%
%

\begin{figure}
    \centering
    \includegraphics[width=1.0\linewidth]{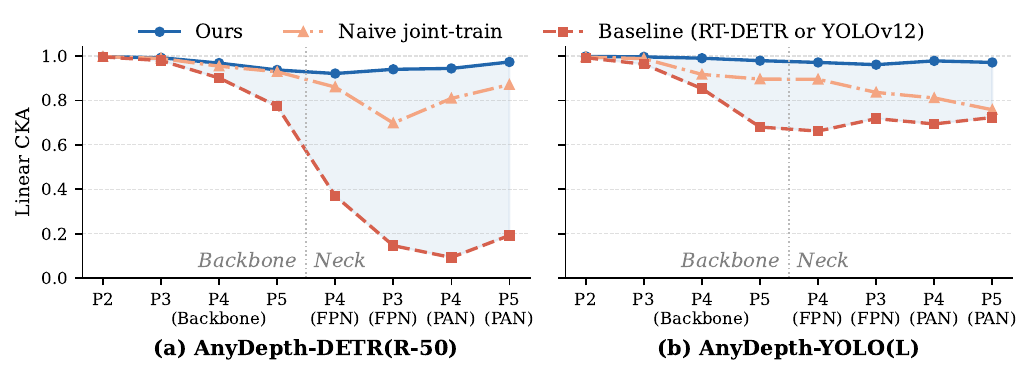}
\caption{
 Linear CKA between essential and full path outputs
         on COCO \texttt{val2017}. 
}
\label{fig:cka}
\end{figure}

Table~\ref{tab:feature_similarity} and Figure~\ref{fig:cka} show that
our any-depth models maintain consistently high alignment
(CKA: 0.92--0.99), while standard baselines exhibit
substantially lower similarity (CKA: 0.09--0.99). 
The naive joint-train, which has only the architectural decomposition
falls between the two extremes with moderate alignment (CKA: 0.70--0.99).
This result confirms that stage-wise modularity requires both the
architectural decomposition, which provides a structural inductive bias
through additive refinement, and self-distillation, which prevents
$\Delta(x_{\text{ess}})$ in Eq.~\ref{eq:residual_stage} from drifting.

In standard RT-DETR, alignment degrades at deeper stages: CKA drops from
0.996 at P2 to 0.774 at P5 and collapses to 0.09--0.37 in the neck. The
naive joint-train improves this (P5: 0.929, neck: 0.69--0.87), while
AnyDepth-DETR maintains CKA $\geq 0.92$ throughout. The gap is larger for
RT-DETR than YOLOv12, as RT-DETR's residual stages accumulate larger
per-block transformations than YOLOv12's CSP stages. The small residual
dissimilarity in our models (CKA gap from 1.0: 0.01--0.08) reflects the
spatial refinements the refinement path is designed to introduce,
consistent with the localization precision gains at strict IoU thresholds.

\begin{table}[t]
\centering
\small
\caption{Linear CKA between base-net and super-net feature representations
         at each backbone and neck stage, measured on COCO \texttt{val2017}.
         Each feature map of shape $(B, C, H, W)$ is spatially reduced via $4{\times}4$ adaptive average pooling and then flattened to $(B{\cdot}16,\, C)$, treating each pooled spatial cell as an independent sample. Features are accumulated up to 20{,}000 spatial samples per stage across 500 batches
         (95\% CIs ${\leq}0.012$, bootstrapped with $n{=}500$).
         \textit{Naive joint}: our architecture without switchable BN/aggregator or self-distillation.
         \textit{Baseline}: standard models (RT-DETR or YOLOv12).
         }
\label{tab:feature_similarity}
\renewcommand{\arraystretch}{1.15}
\setlength{\tabcolsep}{3.5pt}
\begin{tabular}{ll ccc ccc}
\toprule
\multirow{2}{*}{\textbf{Comp.}} &
\multirow{2}{*}{\textbf{Stage}} &
\multicolumn{3}{c}{\textbf{AnyDepth-DETR (R-50)}} &
\multicolumn{3}{c}{\textbf{AnyDepth-YOLO (L)}} \\
\cmidrule(lr){3-5} \cmidrule(lr){6-8}
& & Ours & Naive joint & RT-DETR
  & Ours & Naive joint& YOLOv12 \\
\midrule
\multirow{4}{*}{Backbone}
  & P2 & $0.997$ & $0.997$ & $0.996$ & $0.998$ & $0.995$ & $0.992$ \\
  & P3 & $0.992$ & $0.989$ & $0.979$ & $0.996$ & $0.988$ & $0.963$ \\
  & P4 & $0.968$ & $0.954$ & $0.901$ & $0.990$ & $0.917$ & $0.853$ \\
  & P5 & $0.937$ & $0.929$ & $0.774$ & $0.979$ & $0.896$ & $0.680$ \\
\midrule
\multirow{4}{*}{Neck}
  & P4 (FPN) & $0.921$ & $0.860$ & $0.370$ & $0.971$ & $0.895$ & $0.662$ \\
  & P3 (FPN) & $0.940$ & $0.698$ & $0.146$ & $0.961$ & $0.836$ & $0.718$ \\
  & P4 (PAN) & $0.944$ & $0.809$ & $0.093$ & $0.978$ & $0.811$ & $0.694$ \\
  & P5 (PAN) & $0.973$ & $0.871$ & $0.192$ & $0.971$ & $0.758$ & $0.723$ \\
\bottomrule
\end{tabular}
\end{table}

\begin{figure}
    \centering
    \includegraphics[width=1.0\linewidth]{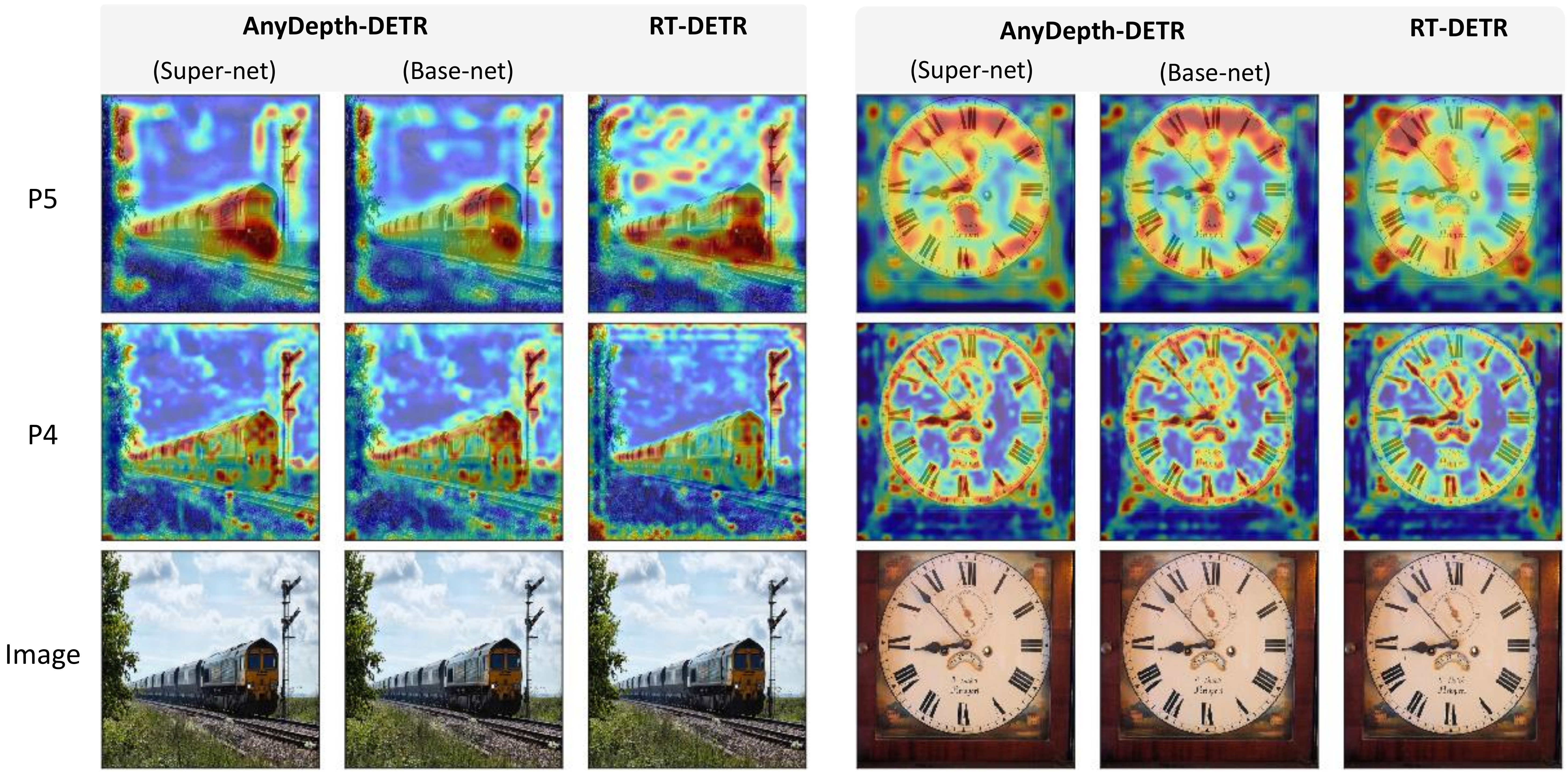}
\caption{
    Activation heatmaps at the P4 and P5 backbone stages. 
}
\label{fig:activation_heatmaps}
\end{figure}

Figure~\ref{fig:activation_heatmaps} provides qualitative support for stage-wise
modularity by visualizing backbone activation heatmaps at stages P4 and P5. 
The super-net and base-net of AnyDepth-DETR(R-50) produce 
similar activation patterns at both stages, attending to the same
semantic regions of the image, despite the base-net executing only the essential path. 

\clearpage

\paragraph{Spatial vs.\ Semantic Feature Alignment.}
Section~\ref{sec:feat_kd} motivates the GAP-based alignment by arguing
that per-location spatial matching would over-constrain the full path.
We verify this empirically by comparing against a \emph{spatial}
alignment variant that matches $\ell_2$-normalized features at each
spatial location:
\begin{equation}
\label{eq:feat_kd_spatial}
\mathcal{L}^{\text{kd}}_{\text{feat,spatial}} =
  \frac{1}{|\mathcal{I}|}\sum_{s \in \mathcal{I}}
  \frac{1}{H_s W_s}\sum_{u=1}^{H_s W_s}
  \left\| \hat{\mathbf{x}}^{\text{ess}}_{s,u} -
          \hat{\mathbf{x}}^{\text{full}}_{s,u} \right\|_2^2,
\qquad
\hat{\mathbf{x}}_{s,u} =
  \frac{\mathbf{x}_{s,u}}{\|\mathbf{x}_{s,u}\|_2}.
\end{equation}
As shown in Table~\ref{tab:feat_alignment_comparison}, spatial alignment
underperforms the GAP-based semantic alignment on both the super-net and
base-net, confirming that per-location matching penalizes the spatial
refinements the refinement path is designed to learn. 
The GAP-based
formulation avoids this by aligning only channel-level descriptors,
preserving the refinement path's freedom to develop richer spatial
representations.

\begin{table}[htb]
\centering
\caption{Comparison of feature alignment strategies for
         AnyDepth-DETR (R-50) on COCO \texttt{val2017}.
         Spatial: per-location $\ell_2$-normalized feature matching;
         Semantic (Ours): GAP-based channel descriptor alignment
         (Eq.~\ref{eq:feat_kd}).}
\label{tab:feat_alignment_comparison}
\renewcommand{\arraystretch}{1.15}
\small
\begin{tabular}{l cc}
\toprule
\textbf{Alignment} & AP$_\text{super}$ & AP$_\text{base}$ \\
\midrule
Spatial            & 53.1 & 50.3 \\
Semantic (Ours)    & \textbf{53.3} & \textbf{50.6} \\
\bottomrule
\end{tabular}
\end{table}



\subsection{Further Evaluation Results}
\label{sec:appendix-eval}

\paragraph{Performance Across Depth Configurations.}
Table~\ref{tab:adn_yolo_ablation} reports
the full component-wise depth sweeps for AnyDepth-YOLO (L).
For AnyDepth-YOLO (L), applying essential paths progressively to
backbone stages reduces GFLOPs from 83.3 to 70.8 (15\%) with only
1.0 AP cost. Applying essential paths to both FPN and PAN neck stages
further reduces GFLOPs to 72.1 at an additional 0.8 AP cost, though
with a smaller FPS gain than the backbone, reflecting the neck's
smaller computational share. Overall, the base-net achieves a
$1.57\times$ speedup over the super-net at only 1.8 AP cost, all from
a single set of weights.

\begin{table}[h]
\centering
\caption{Component-wise ablation of AnyDepth-YOLO (L) on COCO \texttt{val2017}.}
\label{tab:adn_yolo_ablation}
\small
\setlength{\tabcolsep}{6pt}
\renewcommand{\arraystretch}{1.15}
\begin{tabular}{l l r r c}
\toprule
\textbf{Component} & \textbf{Setting} & \textbf{GFLOPs} & \textbf{FPS} & \textbf{AP} \\
\midrule
\rowcolor{gray!12}
\multicolumn{2}{l}{\textbf{Super-net}: All full paths in backbone and neck} & 83.3 & 101.5 & \textbf{53.9} \\
\midrule
\multirow{4}{*}{\shortstack[l]{Backbone}}
 & Ess. path in P2 & 81.6 & 112.6 & 53.7 \\
 & Ess. paths in P2, P3 & 74.7 & 130.2 & 53.3 \\
 & Ess. paths in P2, P3, P4 & 72.8 & 134.2 & 53.0 \\
 & Ess. paths in P2, P3, P4, P5 & 70.8 & 134.1 & 52.9 \\
\midrule
\multirow{2}{*}{\shortstack[l]{Neck\\{\small(FPN+PAN)}}}
 & Ess. paths in PAN & 79.2 & 103.3 & 53.6 \\
 & Ess. paths in FPN+PAN & 72.1 & 113.9 & 52.8 \\
\midrule
\rowcolor{gray!12}
\multicolumn{2}{l}{\textbf{Base-net}: All essential paths in backbone and neck} & 59.6 & 155.3 & \textbf{52.1} \\
\bottomrule
\end{tabular}
\end{table}





\end{document}